\documentclass[10pt,twocolumn,letterpaper]{article}

\usepackage{wifs}
\usepackage{times}
\usepackage{epsfig}
\usepackage{graphicx}
\usepackage{amsmath}
\usepackage{amssymb}
\usepackage{caption}
\usepackage{subcaption}

\usepackage[pagebackref=true,breaklinks=true,letterpaper=true,colorlinks,bookmarks=false]{hyperref}
\wifsfinalcopy 

\def\wifsPaperID{****} 

\ifwifsfinal\fi

\begin{document}

\title{ Gender Effect on Face Recognition for a Large Longitudinal Database}

\author{Caroline Werther$^1$\\
University of North Carolina Wilmington\\
Wilmington, NC\\
{\tt\small cew4278@uncw.edu}
\and
Second Author\\
Institution2\\
First line of institution2 address\\
{\tt\small secondauthor@i2.org}
}

\author{
Caroline Werther$^1$, Morgan Ferguson$^2$, Kevin Park$^3$, Troy Kling$^1$, Cuixian Chen$^1$, Yishi Wang$^1$\\
$^1$University of North Carolina Wilmington, $^2$Elon University, $^3$Emory University\\
{\tt\small{ $^1$\{cew4278, tpk7509, chenc, wangy\}@uncw.edu}, $^2$mferguson11@elon.edu, $^3$kevin.park@emory.edu } }

\maketitle
\thispagestyle{empty}

\begin{abstract}
Aging or gender variation can affect the face recognition performance dramatically. While most of the face recognition studies are focused on the variation of pose, illumination and expression, it is important to consider the influence of gender effect and how to design an effective matching framework. In this paper, we address these problems on a very large longitudinal database MORPH-II which contains 55,134 face images of 13,617 individuals. First, we consider four comprehensive experiments with different combination of gender distribution and subset size, including: 1) equal gender distribution; 2) a large highly unbalanced gender distribution; 3) consider different gender combinations, such as male only, female only, or mixed gender; and 4) the effect of subset size in terms of number of individuals. Second, we consider eight nearest neighbor distance metrics and also Support Vector Machine (SVM) for classifiers and test the effect of different classifiers. Last, we consider different fusion techniques for an effective matching framework to improve the recognition performance.
\end{abstract}

\section{Introduction}
Human faces are important, as the face is the non-verbal portal of a person. Therefore, face authentication has attracted great attentions in both research communities and industries recently, due to its significant role in human computer interaction (HCI), internet access control, security control and surveillance, electronic customer relationship, and health science. Face recognition is widely studied over past few decades. While face recognition algorithms have improved significantly in recent years, much of the previous research is focused on the variation of pose, illumination, expression (PIE) or occlusion. However, the variation of aging and gender may impose significant challenges on the face recognition task, comparing to the variation in PIE.

\begin{figure}[htp]
\centering
\begin{tabular}{ccccc}
\centering
\includegraphics[width = 0.4in,keepaspectratio]{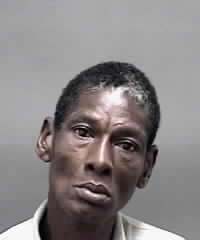} &
\includegraphics[width = 0.4in,keepaspectratio]{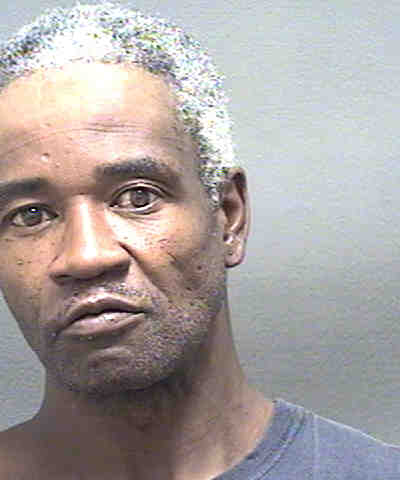} &
\includegraphics[width = 0.4in,keepaspectratio]{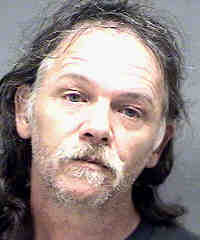}&
\includegraphics[width = 0.4in,keepaspectratio]{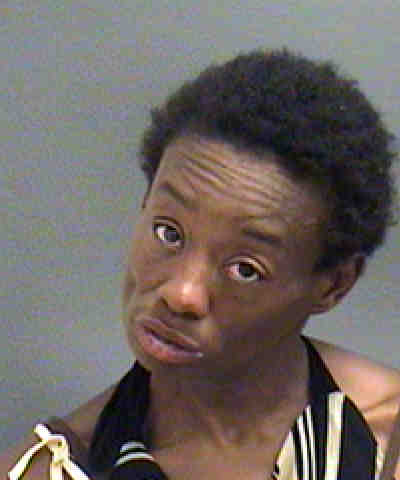} &
\includegraphics[width = 0.4in,keepaspectratio]{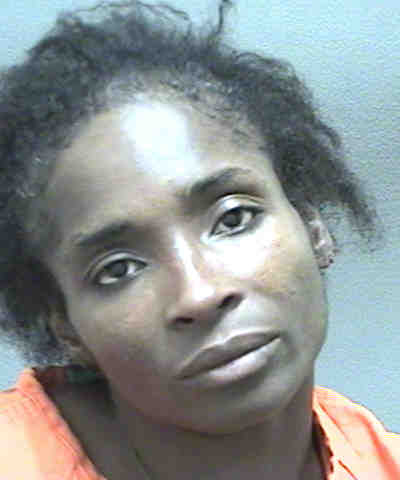}\\
\vspace{0.3cm}
\includegraphics[width = 0.4in,keepaspectratio]{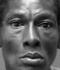} &
\includegraphics[width = 0.4in,keepaspectratio]{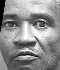} &
\includegraphics[width = 0.4in,keepaspectratio]{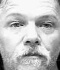}&
\includegraphics[width = 0.4in,keepaspectratio]{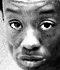} &
\includegraphics[width = 0.4in,keepaspectratio]{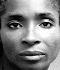}\\
\end{tabular}
\caption{Examples of MORPH-II images before and after preprocessing.}
\label{preprocessing}
\end{figure}

Face recognition with aging variation is refereed to the task of face recognition based on the elapsed time between enrolled and query face images. Comparing to the large amount of studies on face recognition focusing on pose, illumination, and expression, there are relatively limited studies on the effect of age variation
\cite{juefei2011investigating,li2011discriminative,guo2010cross,best2018longitudinal}.
On the other hand, gender variation in face recognition is referred to task of face recognition based on various gender distribution in the query face images database, which could be highly skewed. 
Limited research is available for the impact of soft biometric traits in terms of gender and ethnicity on face recognition performance \cite{narang2016gender,akbari2010performance,hwang2009face,ramesha2009advanced}. However, to our best knowledge, there is no research available in terms of quantitatively analyzing the influence of gender distribution on face recognition for a large database, especially when the gender distribution is highly skewed.

For face recognition, Turk et al introduced PCA in their Eigenfaces approach \cite{turk1991face}, while Belhumeur et al proposed to use LDA as Fisherfaces  \cite{belhumeur1997eigenfaces}. Eigenfaces and Fisherfaces are classical techniques for face feature descriptors in
face recognition \cite{toygar2003face,liu2009comparative,melivsek2008support}.
%
%
On the other hand, various distance metrics such as L1, L2, Cosine Angle, and Mahalanobis are analyzed against the techniques of PCA, LDA, and et al. in \cite{satonkarface,delac2005independent}. 
However, limited research is available in terms of distance fusion, by combining various distance metrics to improve recognition accuracy rates. The role of Eigenvector selection and Eigenspace distance measure on PCA is examined in \cite{yambor2002analyzing}. Distance metrics are tested on their own and then combined using sum rule and bagging to see if there is an improvement in accuracy. Applying the various distance metrics on the FERET database resulted in Mahalanobis distance metric performing the best on it's own when compared to City Block, Squared Euclidean, and Angle distance. No significant improvements are found when combining distance metrics when using the sum rule or bagging methods \cite{yambor2002analyzing}.

This paper is aimed to study systematically how gender will affect the face recognition performance on a large longitudinal database. Next, we want to study if different fusion techniques on various distance metrics can improve the recognition performance. Our main contributions are:
\begin{itemize}
  \item Study the face recognition performance verse various balanced or highly skewed gender distribution on a large database.
  \item Study the performance of various distance metrics fusion techniques to investigate whether distance metrics fusion can improve the recognition accuracy.
  \item Two feature descriptors, eight different distance metrics for classification and five weighting schemes for distance fusion are analyzed systematically to evaluate the face recognition performance.
\end{itemize}

This paper is organized as follows. In Section 2, eight distance metrics are introduced, and various distance metrics fusion approaches are proposed. In Section 3, the database is introduced. Systematic experiments are setup to test the influence of gender effects on face recognition, and the effect of subset size. Experimental results and discussion are given in Section 4. Conclusion and future work are provided in Section 5.

\section{Methods}

\subsection{Feature Extraction}
We consider two facial feature descriptors of Eigenfaces and Fisherfaces to explore the combination of 
eight different distance metrics for classification and five weighting schemes for distance fusion to evaluate the face recognition performance.

Principal component analysis (PCA) is a widely used dimension reduction method. PCA searches for directions in the data that have the largest variance and project the data onto it. It is an orthogonal linear transformation that transforms the data to a new coordinate system such that the greatest variance by some projection of the data comes to lie on the first coordinate which is called the first principal component. Each succeeding component has the highest variance possible under the constraint that it is orthogonal to the preceding components \cite{turk1991face}. This results in a lower dimensional representation of the data, that removes some of the ``noisy" directions, the exact goal we are setting out to achieve for face recognition.  PCA emphasizes variation and brings out strong patterns in a data set. PCA can be a great tool when it comes to dimensionality reduction, however it has its limitations in the fact that it is unsupervised and relies on linear assumptions.


While principal component analysis may be the most famous example of dimensionality reduction, it does have its disadvantages, and that is where linear discriminant analysis comes in. Linear discriminant analysis is used to find a linear combination of features that characterizes or separates two or more classes of data. LDA looks at maximizing the following objective:
\begin{equation}
J(\textbf{w})=\frac{w^TS_Bw}{w^TS_ww}, \label{objective}
\end{equation}

\noindent where $S_B$ is the ``between classes scatter matrix"
and $S_W$ is the ``within classes scatter matrix" \cite{welling2005fisher}.
The scatter matrices are proportional to the covariance matrices, making $S_B$ and $S_W$ symmetric, and we could have defined J using these covariance matrices.




\subsection{Distance Metrics}
Let x and y be two $p \times 1$ feature vectors such that $x=(x_1, x_2,...,x_p)^T$ and $y=(y_1, y_2,...,y_p)^T$. Eight different distance metrics will be considered for face recognition, including: (1) Euclidean Distance (L2), (2) City Block Distance/Manhattan distance (CB), (3) Cosine Distance (COS), (4) Mahalanobis Cosine (MC), (5) Bray Curtis Distance (BC), (6) Canberra Distance (CAN), (7) Correlation (CORR), and (8) Chebyshev Distance (CHEB).

\begin{align*}
dist_{EUC} & =\sqrt{\sum_{i=1}^{n} \ (x_i-y_i)^2},  \\
dist_{CB} & =\sum_{i=1}^{n} |x_i-y_i|, \\
dist_{COS} &=1-\frac{<x, y>}{||x||_2||y||_2},  \textrm{ where} \frac{<x,y>}{||x||_2||y||_2} \textrm{ is cosine similarity},\\
dist_{MC} &=\sqrt{(x-y)V^{-1}(x-y)^T}, \textrm{V is a covariance matrix}\\
dist_{BC} &=\frac{\sum_{i=1}^n|x_i-y_i|}{\sum_{i=1}^{n} |x_i+y_i|}, \\
dist_{CAN} &=\sum_{i=1}^{n}\frac{|x_i-y_i|}{|x_i|+|y_i|},\\
dist_{CORR} &=\frac{\sum_{i=1}^{n}(x-\bar{x})(y-\bar{y})}{\sqrt{\sum_{i=1}^{n}(x-\bar{x})^2\sum_{i=1}^n(y-\bar{y})^2}},\\
dist_{CHEB} &=max|x_i-y_i|  \hspace{.3cm} \text{for all i=1,2,...n}.
\end{align*}


Support Vector Machine (SVM) with a radial basis kernel is considered as another classifying technique in this paper. These two parameters of C and gamma were tuned in order to determine the best possible combination of parameters for our facial recognition algorithm using the radial basis function for SVM.

\subsection{Distance Metric Fusion}

Ensemble Learning has been a developing area of research and interest, and has shown that it could be beneficial. To further improve the overall performance, ensemble learning by fusing multiple predictive decisions to make a final decision could be a potential way to get a more robust decision \cite{polikar2006ensemble}. For example, the classifier ensembles with different combination techniques have been widely explored in recent years. These methods have been shown to potentially reduce the error rate in the classification tasks compared to an individual classifier in a broad range of applications. For facial recognition purposes, ensemble learning involves fusing different classifiers (distance metrics) in order to improve recognition accuracy. 

Although strong distance metrics may result in high accuracy rates on their own, distance metric fusion has the potential to strengthen accuracy rates for weaker distance metrics for a more robust classifier. It is possible that combining results from different methods under ensemble learning can sometimes provide an accuracy rate that is better than an individual method alone through ensemble learning. However, in the decision fusion with ensemble-based systems, it is important to consider the diversity of decisions to be fused, with respect to diverse fusion components.


To access the potential benefits of decision fusion, various combinations of distance metrics are examined. Combinations of the best two, three, and four different distance metrics are tested. All are tested using PCA features and LDA features, as well as a 9 to 1, or 5 to 5 training to testing ratio. The three best and three worst distance metrics are also analyzed for all possible weighting schemes using the 9 to 1 training to testing ratio with the Fisherfaces approach. The subset used for this experimentation is from {\it Experiment 1}.

In this paper, first we consider the  Min-Max feature normalization so that the distance metrics are normalized to have a range from 0 to 1. 
Next, the original distance metrics are combined and analyzed and then compared systematically. 
For all combination of distance metrics analyzed,  several weighting schemes are considered. The average, minimum, median, weighted max pooling (WMP), and weighted average are all tested. It is noticed that for combining two distance metrics, the average and median are the same. The average, minimum, and median are pretty straightforward as far as their computation. For the weighted max pooling (WMP), the distance metrics being analyzed are ordered from the least to greatest distance and then given a weight based on the equation $\frac{e^{distance_i}}{\sum_{n=1}^{i}e^{distance_i}},$
\noindent testing all possible distance metrics $i$, where $i$ is the number of distance metrics to be analyzed. The highest weight resulting from this equation is given to the smallest distance down until the smallest weight is given to the largest distance. 
For the the weighted average, suppose the best three distance metrics are $D_1$, $D_2$, and $D_3$ with corresponding weights of $w_1$, $w_2$, and $w_3$, then the weighted average distance is given by $D^* = w_1 D_1 + w_2 D_2 + w_3 D_3$.



\section{Database and Experiments}
The MORPH-II database \cite{ricanek2006morph} contains 55,134 facial images of 13,617 unique individuals. The images are mugshots taken over a 5 year period and include images of individuals that were arrested once or multiple times. The individual’s ages range from 16 to 77 years and the number of images per individual ranges from 1 to 53, with an average around 4. Because of its size, longitudinal span, and large number of subjects, MORPH-II becomes one of the benchmark dataset in the field of computer vision and pattern recognition. It has been used for a variety of face recognition and demographical analysis.

In the preprocessing step, several measures are taken to clean the available data. Due to variations in expression, lighting, and pose, this dataset imposes potential challenges.  Therefore, all images are preprocessed. Each face image is detected and aligned with eyes centered, cropped, and resized to 70x60 pixels. The images are then histogram equalized to account for varying changes in illumination. These preprocessed, gray level images are used for face recognition. Examples are illustrated in Figure \ref{preprocessing}.

\subsection{Experiments}

For our experiments, only those individuals with 10 or more images were adopted. It leads to a subset containing 83 females and 461 males. Among these individuals, 454 are black, 87 as white, and 3 as Hispanic. This subset gives an unequal race and gender breakdown, which is similar to the entirety of the MORPH-II database. Of the individuals in the subset, we randomly selected 10 images for each subject.

{\bf Experiment 1 (E1):}
In order to create a gender-balanced subset while maximizing subset size, Experiment 1 includes all 83 distinct females and a random selection of 83 distinct males of the 461 available. In total, this subset contains 166 subjects with equal gender representation with 1660 total images. From these images, 1 image is randomly selected for each subject as a testing image and the remaining 9 are used as training images. This results in a training set of 1494 images and a testing set of 166 images. Next, in order to increase the difficulty of the face recognition problem on this subset to see the effect it has on accuracy and computational time, we also consider 5 images for each person as training images and 5 as testing. This results in a training set of 830 images and a testing set of 830 images as well.

{\bf Experiment 2 (E2):}
Our second experiment includes all 544 individuals from our original subset, with 83 females and 461 males. For each individual, 5 images were chosen at random for the training set and 5 were used for the testing set. This results in a much difficult problem in face recognition because of the highly unequal gender distribution (ratio of 1:5 for female v.s male) and larger subset size.

{\bf Experiment 3 (E3): }
Experiment 3 is aimed to analyze the effect of gender distribution on the face recognition problem. Three separate subsets were created from the original subset of 544 subjects with 83 females and 461 males. Each subset contains 82 unique individuals with 5 images for training and the remaining 5 for testing. The "Female Only" subset contains 82 unique females out of the available 83. The "Male Only" subset contains a random selection of 82 unique males out of the 461 available. Lastly, the "Mixed Gender" subset contains a random selection of 41 unique females and 41 unique males out of 544 subjects.

{\bf Experiment 4 (E4):}
The purpose of Experiment 4 is to further analyze the effect of subset size in terms of number of subjects (or individuals) on the face recognition problem based on unequal gender distributions. Four separate subsets were created from the original subset of 544 individuals with 5440 images. To control other influencing factors, each subset contains an exact gender ratio of 5 males to each female. Again, 5 images are used for training and the remaining 5 used for testing. The ``120" subset contains a random selection of 120 distinct individuals, with 20 females and 100 males and a total of 1200 images. Similarly, the ``240" subset contains a random selection of 240 individuals, the ``360" subset contains a random selection of 360 individuals, and the ``480" subset contains a random selection of 480 distinct individuals.

\section{Experiment Results}

\subsection{Results for Experiments 1-4}

 \begin{figure*}
 \centering
 \begin{minipage}{0.5\textwidth}
 \centering
 \includegraphics[width= 0.8\textwidth,keepaspectratio]{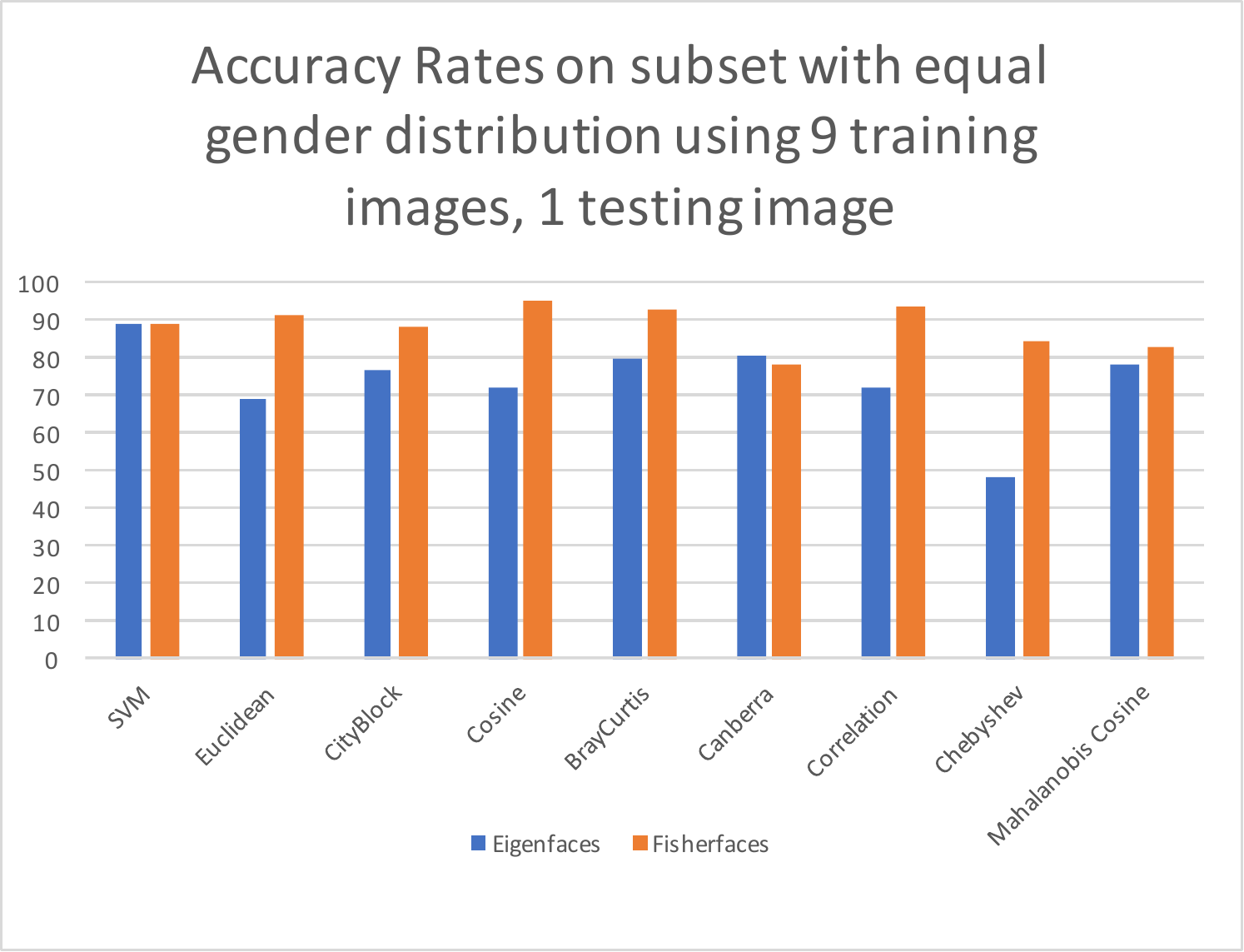}
\caption{{\bf E1} Accuracy for ratio of 9:1 on training:testing.}
\label{e19}
\end{minipage}%
\begin{minipage}{0.5\textwidth}
\centering
\includegraphics[width= 0.8\textwidth,keepaspectratio]{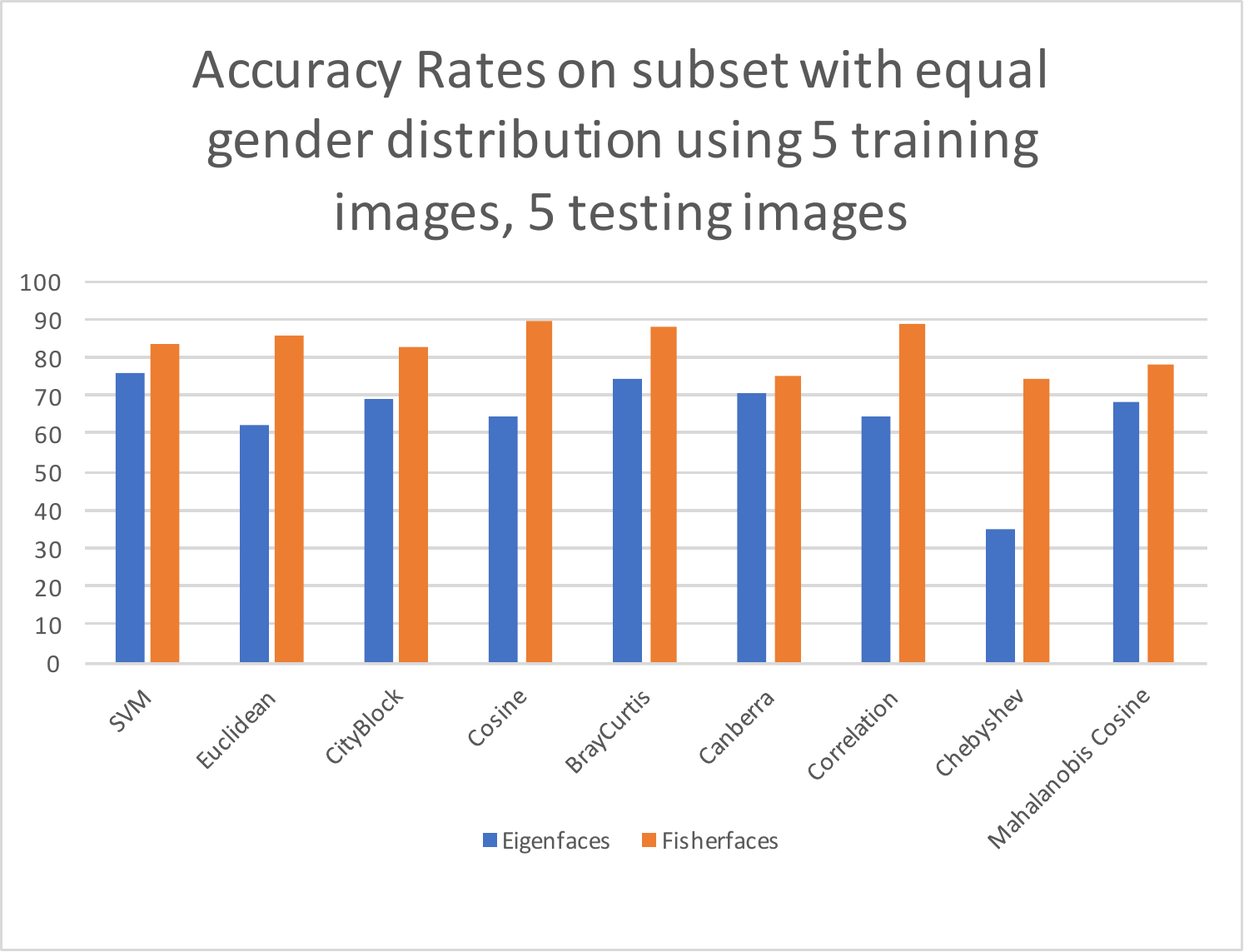}
\caption{{\bf E1} Accuracy for ratio of 5:5 on training:testing.}
\label{e15}
\end{minipage}
\end{figure*}

 \begin{figure*}
 \centering
 \begin{minipage}{0.5\textwidth}
 \centering
 \includegraphics[width= 0.8\textwidth,keepaspectratio]{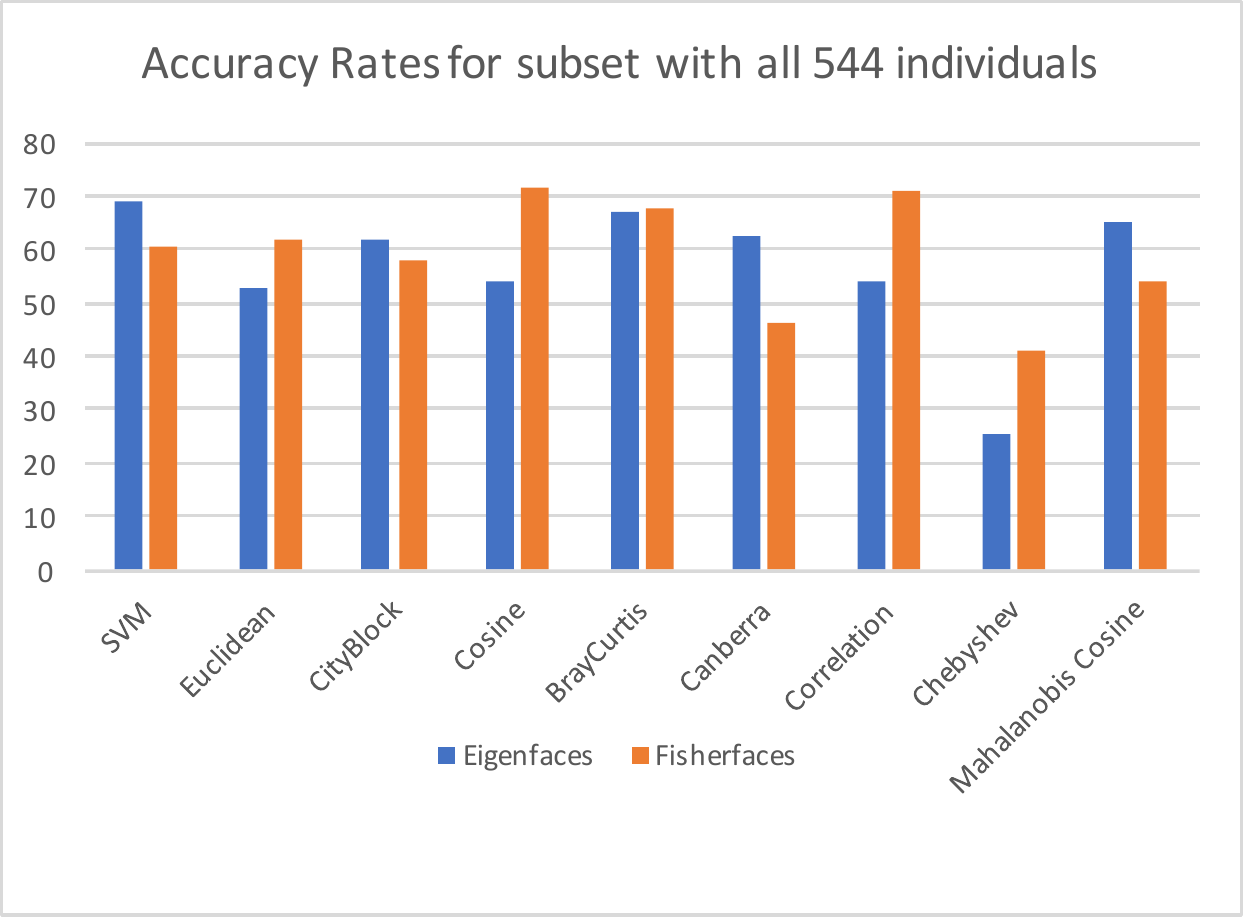}
\caption{{\bf E2} Accuracy for ratio of 5:5 on training:testing.}
\label{e2}
\end{minipage}%
\begin{minipage}{0.5\textwidth}
\centering
\includegraphics[width= 0.9\textwidth,keepaspectratio]{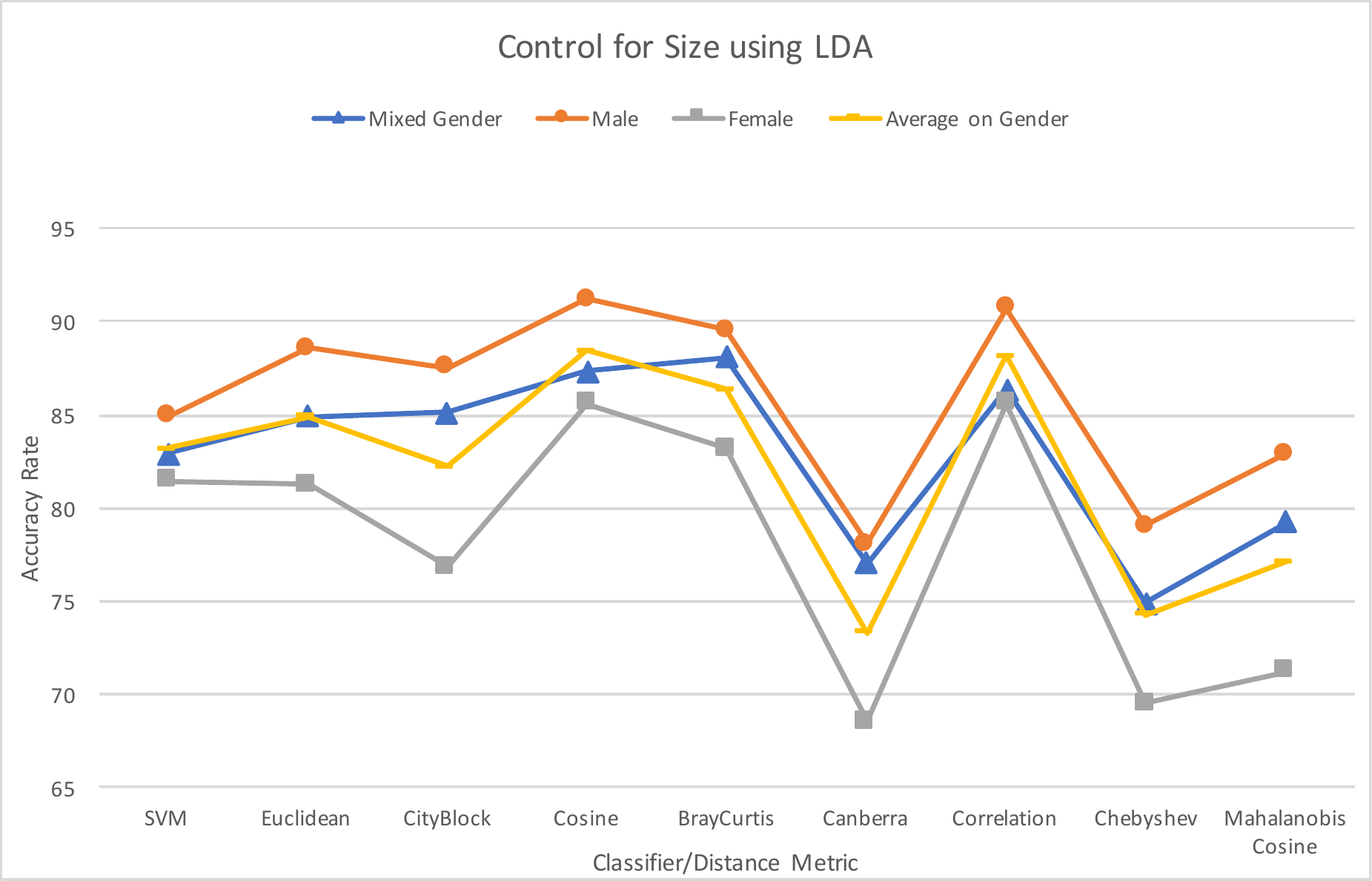}
\caption{{\bf E3} Accuracy for LDA and 5:5 on training:testing.}
\label{e3}
\end{minipage}
\end{figure*}

\begin{figure}[htbp]
\centering
\includegraphics[width = 3in,keepaspectratio]{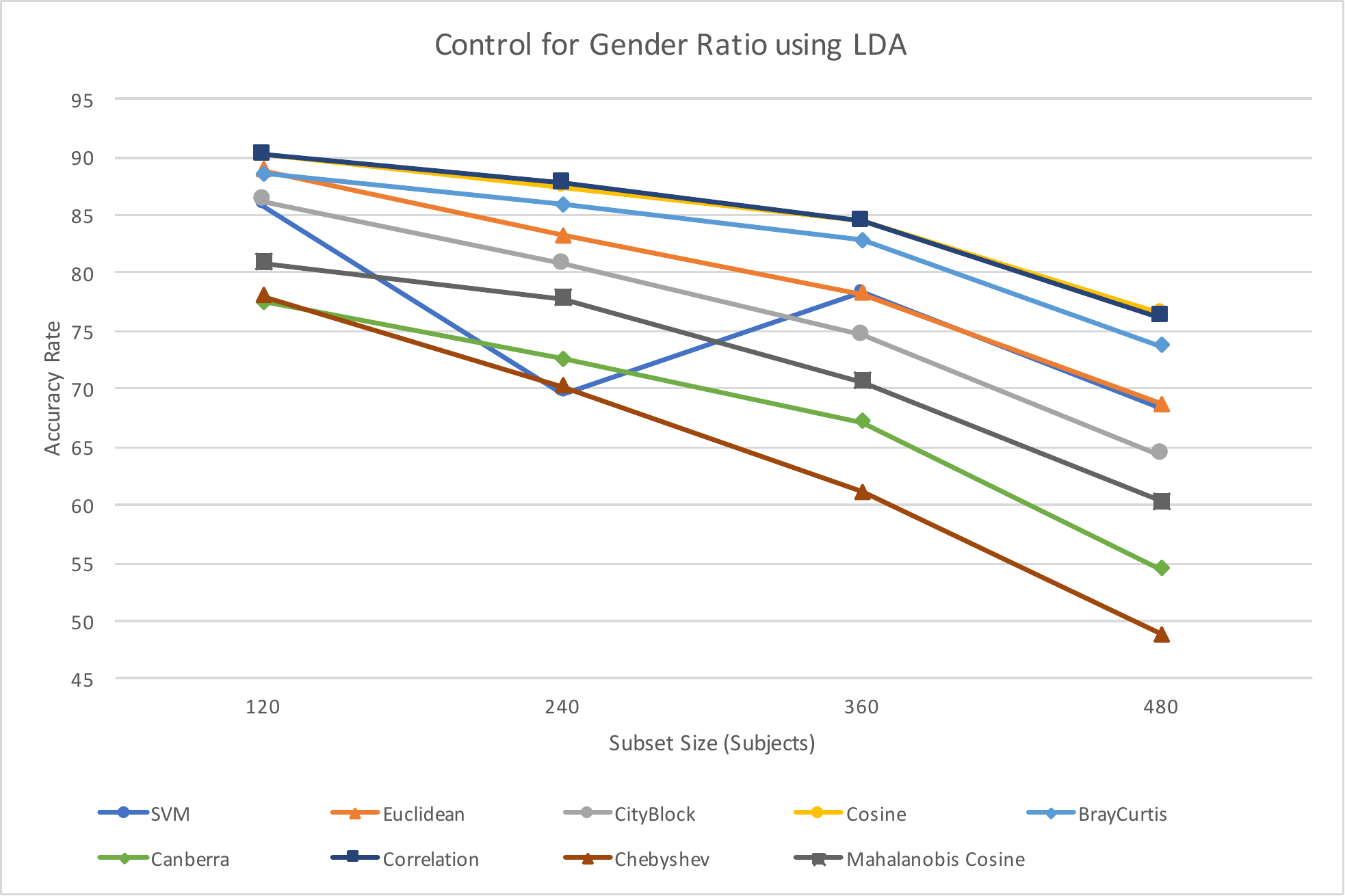}
\caption{{\bf E4} Accuracy for LDA and 5:5 on training:testing.}
\label{e4}
\end{figure}

In this paper, the PCA features (Eigenfaces) and LDA features (Fisherfaces) are considered as the facial representation methods. Our preliminary studies show that the accuracy after the first 100 features for PCA and LDA  keep almost the same. Thus only the first 100 features for both PCA and LDA are adopted in our studies hereafter. Next, the performances on face recognition are obtained by using the various classifying techniques of SVM and eight distance metrics for each experiment with either 9 images for training and 1 image for testing, or a  more difficult problem using 5 images for training and 5 images for testing. 

For Experiment 1 with equal gender distribution subset, Figures \ref{e19} and \ref{e15} show that the accuracy rates are higher with a 9 to 1 training to testing ratio, and overall Fisherfaces yields better results when compared to Eigenfaces for all eight distance metrics. For 9 to 1 training to testing ratio, the combination of PCA and SVM achieves an accuracy of 89.16\%, while the combination of LDA and Cosine distance leads to an accuracy of 95.18\%. On the other hand, for 5 to 5 training to testing ratio, the combination of PCA and SVM achieves 76.27\%, while LDA and Cosine Distance lead to an accuracy of 89.52\%. 

However, for Experiment 2 with highly skewed gender distribution of a 5:1 male to female ratio and a subset size of 544, the accuracy rates are much lower than those of Experiment 1 where the subset size is 166 with equal gender distribution. It is also noticed that the run times increase dramatically from our experiments in Experiment 2, comparing to Experiment 1 \cite{towns2014xsede}.
For 5 to 5 training to testing ratio, the combination of PCA and SVM achieves 68.82\%, while LDA and Cosine Distance lead to an accuracy of 71.47\%. However, further studies are needed to investigate whether the changes in accuracy rates and run time are due to the unequal gender distribution, the increased subset size, or both.

\begin{table}[htbp]
\centering
\resizebox{0.5\textwidth}{!}{%
\begin{tabular}{|c|c|c|c|c|c|c|}
\hline
& \multicolumn{6}{|c|}{5:5 on training:testing} \\ \cline{2-7}
& \multicolumn{2}{|c|}{Male Only} & \multicolumn{2}{|c|}{Female Only} & \multicolumn{2}{|c|}{Mixed Gender} \\ \cline{2-7}
& \multicolumn{2}{|c|}{82 M} & \multicolumn{2}{|c|}{82 F} & \multicolumn{2}{|c|}{41 M, 41 F} \\ \cline{2-7}
& PCA & LDA & PCA & LDA & PCA & LDA \\ \hline
SVM &  \textbf{83.17} & 84.88 & \textbf{72.68} & 81.46 & \textbf{77.07} & 82.93 \\ \hline
Euclidean & 72.44 & 88.54 & 57.07 & 81.22 & 66.34 & 84.88 \\ \hline
CityBlock & 72.20 & 87.56 & 62.44 & 76.83 & 67.56 & 85.12 \\ \hline
Cosine & 73.17 & \textbf{91.22} & 59.51 & \textbf{85.61} & 68.05 & \textbf{87.32} \\ \hline
BrayCurtis & 81.46 & 89.51 & 68.78 & 83.17 & 75.61 & 88.05 \\ \hline
Canberra & 78.29 & 78.05 & 68.05 & 68.54 & 71.22 & 77.07 \\ \hline
Correlation & 73.17 & 90.73 & 59.76 & 85.61 & 67.56 & 86.34 \\ \hline
Chebyshev & 48.29 & 79.02 & 35.12 & 69.51 & 44.63 & 74.88 \\ \hline
Mahal Cos & 70.49 & 82.93 & 60.24 & 71.22 & 65.61 & 79.27 \\ \hline
\end{tabular}
}
\caption{E3: Accuracy Rates (in \%) for LDA features and different gender distributions.}
\label{e3-ar}
\end{table}

\begin{table}[htbp]
\centering
\resizebox{0.5\textwidth}{!}{%
\begin{tabular}{|c|c|c|c|c|c|c|c|c|}
\hline
& \multicolumn{8}{|c|}{5:5 on training:testing} \\ \cline{2-9}
& \multicolumn{2}{|c|}{120 subset} & \multicolumn{2}{|c|}{240 subset} & \multicolumn{2}{|c|}{360 subset} & \multicolumn{2}{|c|}{480 subset} \\ \cline{2-9}
& \multicolumn{2}{|c|}{100 M: 20 F} & \multicolumn{2}{|c|}{200 M: 40 F} &
\multicolumn{2}{|c|}{300 M: 60 F} & 
\multicolumn{2}{|c|}{400 M: 80 F} \\ \cline{2-9}
& PCA & LDA & PCA & LDA & PCA & LDA & PCA & LDA \\ \hline
SVM & \textbf{78.83} & 85.83 & \textbf{74.00} & 69.58 & \textbf{70.89} & 78.28 & \textbf{69.54} & 68.33 \\ \hline
EUC & 66.00 & 88.83 & 57.50 & 83.25 & 55.22 & 78.17 & 53.00 & 68.71	\\ \hline
CB	& 71.17 & 86.17 & 65.08 & 80.75 & 63.61 & 74.61 & 62.29 & 64.33 	\\ \hline
COS & 66.33 & \textbf{90.17} & 59.00 & \textbf{87.33} & 56.50 & \textbf{84.56} & 54.00 & \textbf{76.58} 	\\ \hline
BC	 & 75.33 & 88.50 & 71.92 & 85.92 & 68.61 & 82.78 & 66.67 & 73.67 \\ \hline
CAN & 73.00 & 77.5 & 68.58 & 72.58 & 64.33 & 67.17 & 64.08 & 54.46 	\\ \hline
CORR & 66.00 & 90.17 & 59.00 & 87.75 & 56.67 & 84.44 & 54.00 & 76.17 \\ \hline
CHEB & 42.67 & 78.00 & 33.25 & 70.25 & 30.50 & 61.11 & 28.17 & 48.79 	\\ \hline
MC & 70.00 & 80.83 & 66.67 & 77.75 & 66.39 & 70.67 & 64.96 & 60.29 	\\ \hline
\end{tabular}
}
\caption{E4: Accuracy Rates (in \%) for Face Recognition Algorithms, considering eight distance metrics including: Euclidean Distance (EUC, L2), City Block Distance/Manhattan distance (CB), Cosine Distance (COS), Mahalanobis Cosine (MC), Bray Curtis Distance (BC), Canberra Distance (CAN), Correlation (CORR), and Chebyshev Distance (CHEB).}
\label{e4-ar}
\end{table}

In order to address these two problems,  Experiments 3 and 4 are designed to study the potential impact of the gender effect and also the impact of subset size on accuracy rates and run times individually.  Experiment 3 yields very interesting results. Based on Fisherfaces features, Figure \ref{e3} shows that for LDA features, the male subset has the best accuracy rates consistently for all 8 distance metrics, followed by the mixed gender subset, while the female subset has the worst results. Table \ref{e3-ar} shows that the combination of PCA and SVM and the combination of LDA and Cosine Distance lead to the best accuracies consistently for different gender distributions among Male Only, Female Only, and Mixed Gender of Male and Female.  However, gender appears to have no affect on run time. 

In Experiment 4, for highly skewed gender distribution with a gender ratio of 5 males to 1 female, Table \ref{e4-ar} shows that Cosine Distance and Correlation Distance perform similar on the LDA features which is also illustrated in Figure \ref{e4}, while SVM performs the best on the PCA features. Both Table \ref{e4-ar} and Figure \ref{e4} show that the subset size has a major effect on both accuracy and run time (shown in Appendix). As the subset size increases, accuracy rates decrease and computational times increase.
Overall, our experiment results indicate that for Experiments 1 to 4, SVM works the best method for Eigenfaces, while Cosine works the best for Fisherfaces. Chebyshev is consistently the worst method for both facial features. Consistently, Fisherfaces also yields better accuracy rates than Eigenfaces.

\subsection{Results for Decision Fusion}

We consider eight distance metrics in. Tables \ref{df-best2}, \ref{df-best3}, and \ref{df-best4}, with numbering of (1) Euclidean Distance (EUC, L2), (2) City Block Distance/Manhattan distance (CB), (3) Cosine Distance (COS), (4) Mahalanobis Cosine (MC), (5) Bray Curtis Distance (BC), (6) Canberra Distance (CAN), (7) Correlation (CORR), and (8) Chebyshev Distance (CHEB). To avoid confounding with the gender effect, only the subset from {\it Experiment 1} with equal gender distribution is considered hereafter.
Table \ref{df-best2} shows that, when fusing the best two PCA features, the WMP approach outperforms individual distance metrics for PCA features, while there is no increase in accuracy for LDA features. Tables \ref{df-best3} and \ref{df-best4} show that, when fusing the best three or four PCA features, the median approach outperforms individual distance metrics for PCA features, while either median or weighted average outperform individual distance metrics for LDA features. In summary, the experiment results from Tables \ref{df-best2}, \ref{df-best3} and \ref{df-best4} suggest that distance metrics fusion could have potential benefits for accuracy rates when testing the MORPH-II database.

\begin{table}[htp]
\centering
\begin{tabular}{|c|c|c|c|c|}
\hline
& \multicolumn{4}{|c|}{Best 2} \\ \cline{2-5}
& \multicolumn{2}{|c|}{PCA} & \multicolumn{2}{|c|}{LDA}  \\ \cline{2-5}
& 9:1 & 5:5 & 9:1 & 5:5 \\ \hline
Med (Avg) & 81.33 & 73.37 & 91.57 & 89.04 \\ \hline
Min &  79.52 & 70.12 & 91.57 & 89.04 \\ \hline
WMP & \textbf{81.93} & \textbf{73.37}& \textbf{91.57} & \textbf{89.04} \\ \hline
(0.9, 0.1) & 81.33 & 73.25 & 91.57 & 89.04   \\ \hline
(0.1, 0.9) & 81.33 & 72.53 & 91.57 & 89.04 \\ \hline \hline
Metric-1 & 80.72$^6$ &  72.29$^5$ &  91.57$^7$ & 89.04$^7$  \\ \hline
Metric-2 & 80.12$^4$ &  68.80$^4$ &  91.57$^3$ & 88.92$^3$  \\ \hline
\end{tabular}
\caption{Distance metrics fusion: accuracy rates (in \%) for combining best two normalized distance metrics. The superscripts are of numbering of (1) Euclidean (EUC, L2), (2) City Block/Manhattan (CB), (3) Cosine (COS), (4) Mahalanobis Cosine (MC), (5) Bray Curtis (BC), (6) Canberra (CAN), (7) Correlation (CORR), and (8) Chebyshev (CHEB). Additionally, (0.9, 0.1) are the weights for weighted average approach.}
\label{df-best2}

\bigskip

\begin{tabular}{|c|c|c|c|c|}
\hline
& \multicolumn{4}{|c|}{Best 3}  \\ \cline{2-5}
& \multicolumn{2}{|c|}{PCA} & \multicolumn{2}{|c|}{LDA}  \\ \cline{2-5}
& 9:1 & 5:5 & 9:1 & 5:5  \\ \hline
Avg & 81.33 & 72.77 & 91.57 & 88.92 \\ \hline
Min &  78.31 & 68.19 & 91.57 & 86.75  \\ \hline
Med & \textbf{81.33} & \textbf{74.10} & 91.57 & 89.03  \\ \hline
WMP & 81.33 & 72.77 & 91.57 & 88.92  \\ \hline
(0.8, 0.1, 0.1) & 81.33 & 74.10 & 91.57 & \textbf{89.16}    \\ \hline
(0.4, 0.3, 0.3) & 81.33 & 72.89 & 91.57 & 88.92  \\ \hline
(0.1, 0.1, 0.8) & 80.12 & 71.45 & \textbf{92.77} & 88.80 \\ \hline \hline
Metric-1 & 80.72$^6$ & 72.29$^5$ & 91.57$^3$ & 89.04$^7$ \\ \hline
Metric-2 & 80.12$^4$ & 68.80$^4$ & 91.57$^5$ & 88.92$^3$ \\ \hline
Metric-3 & 78.92$^5$ & 68.31$^6$ & 91.57$^7$ & 86.27$^5$  \\ \hline
\end{tabular}
\caption{Distance metrics fusion: accuracy rates (in \%) for combining best  three normalized distance metrics. }
\label{df-best3}

\bigskip

\resizebox{0.47\textwidth}{!}{%
\begin{tabular}{|c|c|c|c|c|}
\hline
& \multicolumn{4}{|c|}{Best 4}  \\ \cline{2-5}
& \multicolumn{2}{|c|}{PCA} & \multicolumn{2}{|c|}{LDA}  \\ \cline{2-5}
& 9:1 & 5:5 & 9:1 & 5:5 \\ \hline
Avg & 80.72 & 73.13 & 92.77 & 88.92  \\ \hline
Min & 75.30 & 66.87 & 90.36 & 84.34   \\ \hline
Med & \textbf{81.33} & \textbf{73.25} & 92.17 & \textbf{89.52}  \\ \hline
WMP & 80.72 & 73.25 & 92.77 & 88.92   \\ \hline
(0.4, 0.4, 0.1, 0.1) & 81.33 & 73.25 & 92.17 & 88.80   \\ \hline
(0.3, 0.3, 0.2, 0.2) & 80.72 & 73.37 & 92.77 & 88.92  \\ \hline
(0.1, 0.1, 0.4, 0.4) & 80.72 & 72.65 & \textbf{92.77} & 88.92   \\ \hline\hline
Metric-1 & 80.72$^6$ & 72.29$^5$ & 91.57$^3$ & 89.04$^7$ \\ \hline
Metric-2 & 80.12$^4$ & 68.80$^4$ & 91.57$^5$ & 88.92$^3$   \\ \hline
Metric-3 & 78.92$^5$ & 68.31$^6$ & 91.57$^7$ & 86.27$^5$    \\ \hline
Metric-4 & 76.51$^2$ & 67.35$^2$ & 86.75$^1$ & 80.72$^1$  \\ \hline
\end{tabular}
}
\caption{Distance metrics fusion: accuracy rates (in \%) for combining best four normalized distance metrics. }
\label{df-best4}
\end{table}

\section{Conclusion}

When looking at testing single classifiers, SVM is the best technique for Eigenfaces and Cosine Distance is the best for Fisherfaces. Fisherfaces provide better accuracy rates and do not prove to be any more computationally intensive than Eigenfaces. A training to testing ratio of 9 to 1 provided better accuracy rates and is less computationally intensive than a ratio of 5 to 5. Overall, the combination of Fisherfaces and the Cosine Distance metric work the best for MORPH-II data in our study. 
Based on Fisherfaces features, interesting findings show that gender can have significant impact on face recognition that the male subset has the best accuracy rates, followed by the mixed gender subset, while the female subset has the worst results. However, gender appears to have no affect on run time. On the other hand,  it is showed that the subset size can have a major effect on both accuracy and run time. As the subset size increases, accuracy rates decrease and computational times increase. We hope these findings can shed some lights on the problem of gender effect on face recognition.
When analyzing the distance metrics fusion, while it remains unclear what weighting scheme may be the best to use, overall the accuracy increases or stays the same, regardless of the weighting scheme.
%
%
Future study includes further investigation for gender effect on face recognition under Deep Learning framework can be considered.

\section{Acknowledgements}

This material is based in part upon work supported by the NSF
 DMS-1659288. 
%
This work used the XSEDE Stampede2 at the TACC through allocation DMS170019. 

{\small
\bibliographystyle{ieee}
\bibliography{referencest}
}

\end{document}